\documentclass{edm_template}
\usepackage{booktabs}
\usepackage{comment}
\usepackage{url}

\usepackage{graphicx}
\usepackage{svg}
\usepackage{multicol}

\newcommand{\R}{\mathbb{R}}

\newcommand{\thetastraj}{\vec{\theta}_{s,t}}

\newcommand{\Xstvec}{\vec{x}_{s,t}}
\newcommand{\Ystvec}{\vec{y}_{s,t}}

\begin{document}

\title{Back to the basics: Bayesian extensions of IRT outperform neural networks for proficiency estimation}


\numberofauthors{1} %  in this sample file, there are a *total*
\author{
% 1st. author
\alignauthor
Kevin H.~Wilson\titlenote{Contributed equally to the work.},
Yan Karklin\raisebox{9pt}{$\ast$},
Bojian Han\titlenote{Performed initial coding and analysis while at Knewton.},
Chaitanya Ekanadham\raisebox{9pt}{$\ast$}
\\
\affaddr{$^\ast$Knewton, Inc. New York, NY \hspace{2em} 
$^\dagger$Carnegie Mellon University. Pittsburgh, PA}\\
\email{\{kevin,yan,chaitu\}@knewton.com \hspace{3em} bojianh@andrew.cmu.edu}
}

\maketitle
\begin{abstract}
Estimating student proficiency is an important task for computer based learning systems. We compare a family of IRT-based proficiency estimation methods to Deep Knowledge Tracing (DKT), a recently proposed recurrent neural network model with promising initial results. We evaluate how well each model predicts a student's future response given previous responses using two publicly available and one proprietary data set. We find that IRT-based methods consistently matched or outperformed DKT across all data sets at the finest level of content granularity that was tractable for them to be trained on. A hierarchical extension of IRT that captured item grouping structure performed best overall. When data sets included non-trivial autocorrelations in student response patterns, a temporal extension of IRT improved performance over standard IRT while the RNN-based method did not. We conclude that IRT-based models provide a simpler, better-performing alternative to existing RNN-based models of student interaction data while also affording more interpretability and guarantees due to their formulation as Bayesian probabilistic models.  
\end{abstract}

\section*{Acknowledgements}

Many thanks to Siddharth Reddy, David Kuntz, Kyle Hausmann, and Celia Alicata for discussions of this work and help editing the manuscript.

 % NOT required for Proceedings

\section{Introduction}\label{sec:intro}

% Why do we care about proficiency estimation?

A key challenge for computer-based learning systems is to estimate a student's proficiency based on her previous interactions with the system. Accurate estimation of proficiency enables more efficient diagnosis and remediation of her weaknesses and more effective advancement of her knowledge frontier. Proficiency estimates can also provide the student or teacher with actionable information to improve student outcomes when reported as analytics \cite{KnewtonWhitePaper}.

% What's the state-of-the-art and what are the open questions?

Two classical families of methods for estimating proficiency are Item Response Theory (IRT) \cite{IRTGeneral, Lord1952} and Bayesian Knowledge Tracing (BKT) \cite{Corbett95}. IRT essentially amounts to structured logistic regression (see Section~\ref{sec:irt}), estimating latent quantities corresponding to student ability and assessment properties such as difficulty. BKT does not capture assessment properties but employs a {\em dynamic} representation of student ability. A growing body of recent work has focused on modeling various structural properties of students and assessments in an attempt to combine the advantages of IRT and BKT, for instance \cite{PardosHeffernan10, PardosHeffernan11, SPARFA14, FAST14, Khajah14, LeeBrunskill12, EkanadhamKarklin15}). In a recently proposed method known as Deep Knowledge Tracing (DKT) \cite{DKTNIPS}, a recurrent neural network was trained to predict student responses and was shown to outperform the best published results (\cite{PardosHeffernan11}) on the publicly available ASSISTments data set \cite{AssistmentsDataSet} by about 20 percentage points with respect to the AUC metric described in Section~\ref{sec:evaluation}. 

% What did we do and what did we find?

To investigate DKT's advantage over traditional models, we compared a standard one parameter IRT model, two extensions of that model, and DKT on three data sets (two are publicly available and one is proprietary) on a realistic online prediction task that is typically required by computer-based learning systems (see Section~\ref{sec:evaluation}), and which was consistent with the evaluation task employed in \cite{DKTNIPS}.\footnote{Code for the IRT and DKT models, as well as instructions for reproducing our results, can be found at {\tt github.com/Knewton/edm2016}.}  We reproduce the results of \cite{DKTNIPS} on the ASSISTments data set, but find that proper accounting for duplicate data negates the claimed performance gains. For the two larger data sets, computational tractability hampered our ability to train DKT on fine-grained content labels, while training IRT-based models scaled to handle them. Moreover, the IRT-based models' best tractable performance matches or outperforms DKT's best tractable performance on all data sets, with a hierarchical extension of IRT performing the best in all cases. We conclude that for these data sets, IRT-based models provide simple, better-performing alternatives to DKT while also affording more interpretability and guarantees due to their formulation as Bayesian probabilistic models.

% Summary of paper organization

%The rest of this paper is organized as follows. Section~\ref{sec:models} describes the models that we will compare: standard IRT (Section~\ref{sec:irt}), a hierarchical extension to account for grouped assessments (Section~\ref{sec:hirt}), an extension of IRT \cite{EkanadhamKarklin15} that allows student proficiencies to fluctuate over time (Section~\ref{sec:tskirt}), and Deep Knowledge Tracing \cite{DKTNIPS} (Section~\ref{sec:dkt}). Section~\ref{sec:datasets} describes the three data sets on which we evaluate these models: the ASSISTments dataset \cite{AssistmentsDataSet} (Section~\ref{sec:assistments}), the KDD Cup 2010 dataset \cite{KDDCupDataSet} (Section~\ref{sec:kddcup}), and a subset of Knewton's proprietary data (Section~\ref{sec:knewtondata}). In Section~\ref{sec:evaluation}, we describe our method for evaluating the models on these datasets. We then describe our findings in Section~\ref{sec:results} and conclude in Section~\ref{sec:conclusion}.

\section{Models of student responses}\label{sec:models}

In this section we set notation and describe the models we compare. Throughout, we will represent the student response data $D$ as a set of tuples $(s,i,r,t)$ indicating the student, item, correctness, and time of each response. In this paper, time will be indexed by interaction index (rather than wall clock time).

\subsection{Item Response Theory (IRT)}\label{sec:irt}

Item Response Theory (IRT) is a standard framework for modeling student responses dating back to  the 1950s \cite{IRTGeneral, Lord1952}. A single number, called the {\em proficiency} or {\em ability},  represents a student's knowledge state during the course of completing several assessments. It is assumed that this proficiency is not changing during this examination.\footnote{For an in depth discussion of IRT and a review of related literature see \cite{RuppDCM}, especially Chapter 5.}

The model assumes that many students have completed a test of dichotomous items and assigns each student $s$ a proficiency $\theta_s \in \R$. A key innovation of IRT is to model variation across different items. In its simplest form, the {\em one-parameter model}, each item $i$ is assigned a parameter $\beta_i$, representing the {\em difficulty} of the item. The probability that a student $s$ answers item $i$ correctly is given by $f(\theta_s - \beta_i)$, where $f$ is some sigmoidal function.

When $f$ is the logistic function, this corresponds to (structured) logistic regression, where the factors for a response to an item are indicators for students and items. We use a variant of this model known as 1PO (one-parameter ogive) IRT, where the link function $f(x) = \Phi(x)$
%\[ f(x) = \int_{-\infty}^x \frac{1}{\sqrt{2\pi}} \exp\left(\frac{-x^2}{2}\right) dx \]
is the cumulative distribution function of the standard normal distribution\footnote{The ogive yields nearly identical results to the commonly used logistic link function, but allows closed-form posterior computation in the temporal IRT model described in Sec.~\ref{sec:tskirt}}. The maximum likelihood solution of $\{\theta_s, \beta_i\}$ is underdetermined \footnote{For example, the response predictions are invariant when adding a constant offset to the $\{\theta_s\}$'s and $\{\beta_i\}$'s.}; we take a Bayesian approach and regularize the solution of $\{\theta_s, \beta_i\}$ by imposing independent standard normal prior distributions over each $\theta_s$ and $\beta_i$.

\subsubsection{Learning} \label{sec:irtlearning}

To train the parameters on student response data, we maximize the log posterior probability of $\{\theta_s, \beta_i\}$ given the response data (the set of response correctnesses $\{r: (s,i,r,t) \in D\}$, each of which is 0 or 1). Assuming independent, standard normal priors on each $\theta_s$, $\beta_i$, the log posterior is:
\begin{align}
&\log P(\{\theta_s\},\{\beta_i\}|D) = \nonumber \\
&\sum_{(s,i,r,t) \in D}r\log f(\theta_s-\beta_i) + (1-r)\log(1-f(\theta_s-\beta_i)) \nonumber \\
&- \frac{1}{2}\sum_s\theta_s^2 - \frac{1}{2}\sum_i\beta_i^2 + C\, .
\end{align}
We maximize this objective with respect to the parameters using standard second-order ascent methods to obtain the maximum a posteriori (MAP) estimate of each parameter.

\subsection{Hierarchical IRT (HIRT)} \label{sec:hirt}

% CE adding some more context for HIRT here
In many situations, including each of our data sets, the assessment items may have structure that can inform predictions of student responses. For example, groups of items may assess the same topic, resulting in item properties that are more similar within groups than across them Alternatively, items may be derived from common templates.  Templates, often found in math courses, look like ``What is $x + y$?'' and a particular instantiation is generated by choosing values for $x$ and $y$. For example, the ASSISTments data set contains several {\em problems}, many of which are 
with the same {\em template}, many of which in turn assess a single {\em skill}.

We can augment the IRT model to incorporate knowledge about item groups, resulting in a hierarchical IRT model (HIRT). Each item $i$ is associated with a group $j(i)$ whose difficulty is distributed normally around a per-group mean $\mu_{j(i)}$: $\beta_i \sim N(\mu_{j(i)}, \sigma^2)$. Each $\mu_j$ is in turn distributed according to the hyperprior $\mu_j \sim N(0, \tau^2)$. This reflects the belief that the difficulty of items in the same group should be similar. The degenerate cases provide some intuition: the limit $\sigma \to 0$ is the same model as 1PO IRT where we consider the items in the group to be the same item, and the limit $\tau \to 0$ is equivalent to a 1PO IRT model with no groupings.

\subsubsection{Learning}

Learning is done similarly to  Bayesian IRT (section \ref{sec:irt}), except that we ascend the {\em modified} log posterior probability
\begin{align}
&\log P(\{\theta_s\},\{\beta_i\},\{\mu_t\}|D) = \nonumber \\ 
&\sum_{(s,i,r,t) \in D}r\log f(\theta_s-\beta_i) + (1-r)\log(1-f(\theta_s-\beta_i)) \nonumber \\ 
&- \frac{1}{2}\sum_s\theta_s^2 - \frac{1}{2\sigma^2}\sum_i(\beta_i- \mu_{j(i)})^2 - \frac{1}{2\tau^2}\sum_j \mu_j^2 + C\, .
\end{align}

We maximize this objective with respect to $\{\theta_s,\beta_i,\mu_j\}$.

\subsection{Temporal IRT (TIRT)}\label{sec:tskirt}

1PO IRT and HIRT assume each student's knowledge state remains constant over time. However, in a setting where a student may be acquiring (or forgetting) knowledge over a period of time (e.g., while interacting with a tutoring system), we can extend this model by modeling each $\theta_s$ as a stochastic process varying over time (see for example \cite{FAST14}). We adopt the approach described in \cite{EkanadhamKarklin15}, modeling the student's knowledge as a Wiener process:

\begin{align}
&P(\theta_{s,t+\tau} | \theta_{s,t})  = e^{-\frac{(\theta_{s,t+\tau} - \theta_{s,t})^2}{2\gamma^2 \tau}} ~\forall s,t,\tau \, . \label{eq:wiener}
\end{align}
In other words, the change in student $s$'s knowledge state between time $t$ and a future time $t+\tau$ (expressed as $\theta_{s,t} - \theta_{s,t+\tau}$) is normally distributed about 0 with variance $\gamma^2\tau$ where $\gamma$ is a parameter controlling the ``smoothness'' with which the knowledge state varies over time.

\subsubsection{Learning}

We fit the parameters according to the procedure described in \cite{EkanadhamKarklin15}. Estimating the entire trajectory $\thetastraj$ for each student simultaneously with item parameters is very expensive and difficult to do in real-time. To simplify the approach, we learn parameters in two stages:
\begin{enumerate}
\item We learn the ${\beta_i}$ according to a standard 1PO IRT model (see Section~\ref{sec:irtlearning}) on the training student population and freeze these during validation.
\item For each response of each student in the held-out validation population, we predict this response according to a temporal IRT model given the student's previous responses, as described below. For further details of the validation procedure, see Section~\ref{sec:evaluation}.
\end{enumerate}
For the second step, we combine the approximation:
\begin{align}
&P(\{(s', i, r, t') \in D: s'=s, t'\leq t\}|\theta_{s,t}) \approx \nonumber \\  &\prod_{(s',i,r,t') \in D: s'=s, t'\leq t} P((s',i,r,t')|\theta_{s,t})
\end{align}
with \eqref{eq:wiener}, integrating out previous proficiencies of the student to get a tractable approximation of the log posterior over the student's current proficiency given previous responses:
\begin{align}
\log P(\theta_{s,t}|D) &\approx   
\sum_{\substack{(s',i,r,t') \in D \\ s'=s,t'\leq t}} [ r\log f(\tilde{\alpha}_{t'} (\theta_{s,t}-\beta_i)) + \nonumber \\
&(1-r)\log(1-f(\tilde{\alpha}_{t'}(\theta_{s,t}-\beta_i))) ]\, ,
\end{align}
where $\tilde{\alpha}_{t'}=\left(1 + \gamma^2(t-t')\right)^{-1/2}$ . The $\tilde{\alpha}_t$'s are essentially discounting the relative effect of older responses when estimating the current proficiency. See \cite{EkanadhamKarklin15} for details.

\subsection{Deep Knowledge Tracing (DKT)}\label{sec:dkt}

Recently, a recurrent neural network
was used to predict student responses \cite{DKTNIPS}.  Such architectures have seen enormous success in applications to a wide range of other domains (e.g., image processing \cite{DRAWNetwork}, speech recognition \cite{hinton2012deep}, and natural language processing \cite{GrammarAsAForeignLanguage}).

In this model, the input vectors are representations of whether the student answered  a particular question correctly or incorrectly at the previous time step, and the output vectors are representations of the probability, over all the questions in the question bank, that a student will get the question correct at the following time step. In \cite{DKTNIPS}, the authors propose using a one-hot vector $\Xstvec \in \R^{2I}$ to represent the response of a student $s$ (on item $i$) at time $t$. Here $I$ is the total number of items and the first $I$ slots represent answering correctly and the remaining $I$ slots represent answering incorrectly. Output vectors $\Ystvec \in \R^I$ are vectors of probabilities, where the $i$th element of $\Ystvec$ is the model's predicted probability that student $s$ would answer item $i$ correctly at time $t + 1$.

We use a model with one hidden layer, of dimension $H$, which is fully connected\footnote{Note that in \cite{DKTNIPS}, an LSTM network was used in addition to the RNN described here, and the performance of the two networks was comparable.} to both the input and output layers, as well as recurrently to itself. This model is able to capture temporal effects (via the recurrent component of the network) and remains flexible enough to describe non-trivial relationships between items.

\subsubsection{Learning and Parameter Choices}

In order to make learning tractable, we reduced the dimensionality of the input by projecting the $\Xstvec \in \R^{2I}$ to a lower dimensional space $\R^C$ using a random projection matrix $c : \R^{2I} \to \R^C$, as was done in \cite{DKTNIPS}.  We used batch gradient ascent with dropout \cite{Dropout}, and chose the input dimensionality $C$ and the hidden dimensionality $H$ by sweeping these parameters on a data set that was held out from the data used for training and cross-validation.

The predictions are given by the following equations:
\begin{align}
  \vec{h}_{s,t+1} & =  g(W_{hh}\vec{h}_{s,t} + W_{xh}c(\Xstvec) + \vec{b}_h) \\
  \vec{y}_{s,t+1} & =  \phi(W_{hy}\vec{h}_{s,t+1} + \vec{b}_y)
\end{align}
Here, $g$ and $\phi$ are the logistic and arctangent functions, respectively. The parameters of the model $W_{hh},W_{xh},W_{hy},\vec{b}_h, \vec{b}_y$ are fit by optimizing the cross-entropy of the responses with the predicted probabilities (which is equivalent to the log likelihood if these probabilities were produced via a generative probabilistic model):
\begin{equation}
\sum_{(s,i,r,t) \in D} r\log y_{s,t,i} + (1-r)\log (1 - y_{s,t,i})
\end{equation}
Stochastic gradient ascent with minibatches of students on the unrolled RNN, coded using Theano \cite{Theano}, was used to optimize this objective function. 

\section{Data sets}\label{sec:datasets}

In order to test these models, we used three data sets, two publicly accessible and one proprietary. Each of these data sets comes from a system in which students interact with a computer-based learning system in a variety of educational settings (e.g., interspersed with classroom lectures, offline work, etc.).

\begin{figure*}[t!]
\includegraphics[width=\textwidth]{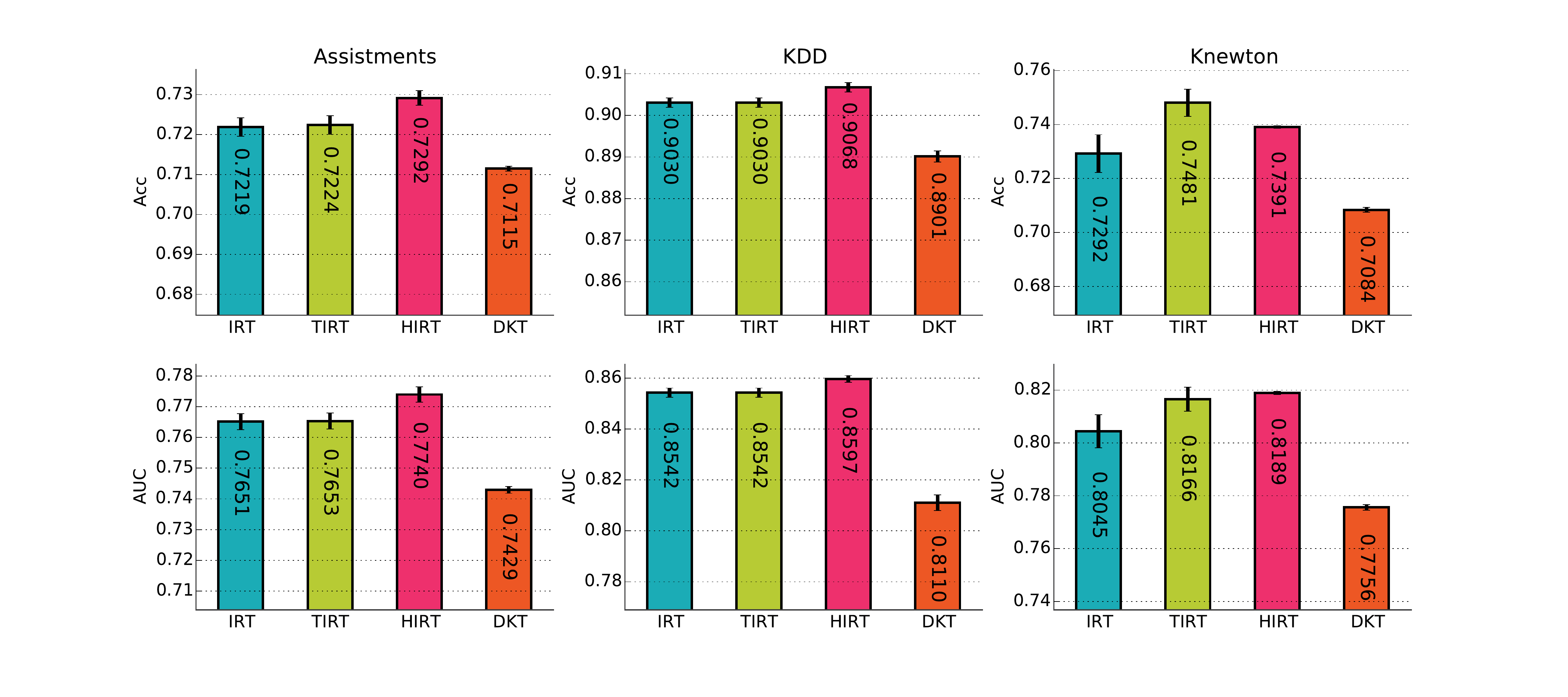}
\vspace{-20pt}
\caption{Summary of results across models and metrics. Error bars represent the standard error of measure of the metric across five folds. For TIRT, parameter selection yielded $\gamma^2 = 0.01$ for ASSISTments, $\gamma^2 = 0$ for KDD (making it identical to IRT), and $\gamma^2 = 100.0$ for Knewton. For HIRT, parameter selection yielded $\sigma^2 = 0.125$ and $\tau^2 = 0.5$ for ASSISTments, $\sigma^2 = 0.5$ and $\tau^2 = 0.25$ for KDD, and $\sigma^2 = 0.25$ and $\tau^2 = 0.125$ for Knewton. For DKT, $C = 50$, $H = 100$, and the probability of dropout is $0.25$ for all models.}\label{fig:results}
\end{figure*}

\subsection{ASSISTments}
\label{sec:assistments}
This data set comes from the ASSISTments product, an online platform which engages students with formative assessments replete with scaffolded hints. Most assessments are templated, and each problem is aligned with one, several, or none of the skills that the product is attempting to teach.

The data set \cite{AssistmentsDataSet} is divided in two parts, the ``skill builder'' set associated with formative assessment and the ``non skill builder'' set associated with summative assessment. All of our results are reported on the ``skill builder'' data set as we expect a stronger temporal signal from formative assessment than from summative assessment. This was also the evaluation data set for \cite{DKTNIPS}.

In preprocessing the data, we associated items not aligned with a skill to a designated ``dummy'' skill, as was done in \cite{DKTNIPS}.   We chose to discard rows duplicating a single interaction (represented by a unique {\tt order\_id} value), a step we do not believe was taken by \cite{DKTNIPS}. These duplicate rows arise when a single interaction is aligned with multiple skills. Without removing these duplicates, models that process all skills simultaneously, including DKT and the IRT variants used in this paper, will see the same student interaction several times in a row, essentially providing these models access to the ground truth when making their predictions. This can artificially boost prediction results by a significant amount (see Section~\ref{sec:results}), as these  ``duplicate'' rows account for approximately 25\% of the rows. Indeed, we observed that the performance gains of DKT are negated when these duplicates are removed (see Section~\ref{sec:results}). Note that typical BKT-based approaches are not susceptible to this artificial boost, since they usually split the data by skill and train separate models. 

After pre-processing, the data set consisted of 346,740 interactions for 4,097 users on 26,684 items arising from 815 templates and 112 skills. The overall percent correct was 64.54\%.

\subsection{KDD Cup} \label{sec:kddcup}
In 2010, the PSLC DataShop released several data sets derived from Carnegie Learning's Cognitive Tutor in (Pre-)Algebra from the years 2005--2009 \cite{KDDCupDataSet}. We used the largest of the ``Development'' data sets, labeled ``Bridge to Algebra 2006--2007.''

One distinct difference between Carnegie Learning's product and ASSISTments is that Carnegie Learning provides much finer representations of the concepts assessed by an individual item. In particular, Carnegie Learning is built around scaffolded, formative assessment, where each {\em step} a student takes to answer a {\em problem} is counted as a separate interaction, with each step potentially assessing different skills (called Knowledge Components (KCs) in the data set). Note that this ``Problem $\to$ Step'' structure provides a hierarchy which HIRT (Section~\ref{sec:hirt}) can exploit.

Like ASSISTments, any particular interaction may assess zero or more skills. We follow the same methodology as we did in Section~\ref{sec:assistments}, arbitrarily but consistently retaining only one of the skills after preprocessing, and associating items not associated with any skills with a designated ``dummy'' skill.

After pre-processing, the data set retained 3,679,198 interactions for 1,146 users on 207,856 steps arising from 19,355 problems and 494 KCs. The overall percent correct  was 88.82\%. 

\subsection{Knewton} \label{sec:knewtondata}
Data was collected from a variety of educational products integrated with Knewton's adaptive learning platform and used in various classroom settings across the world. These products vary with respect to the educational content used (disciplines spanned math, science, and English language learning) as well as the way in which students are guided through the content. For example, students may take an initial assessment and then be remediated on areas needing improvement. In other products, students start from the beginning and  work toward a predefined goal set by the teacher. In all of these settings, Knewton receives data about each interaction (the $(s,i,r,t)$ tuple of Section~\ref{sec:models}). We utilized approximately 1M responses of 6.3K randomly sampled students on 105.6K questions spanning roughly 4 months. Students who worked on fewer than 5 questions total were excluded. After pre-processing, student history lengths ranged from 5 to 3.2K responses. The overall percent correct of these responses is 54.6\%.

\begin{figure*}[t!]
\includegraphics[width=\textwidth]{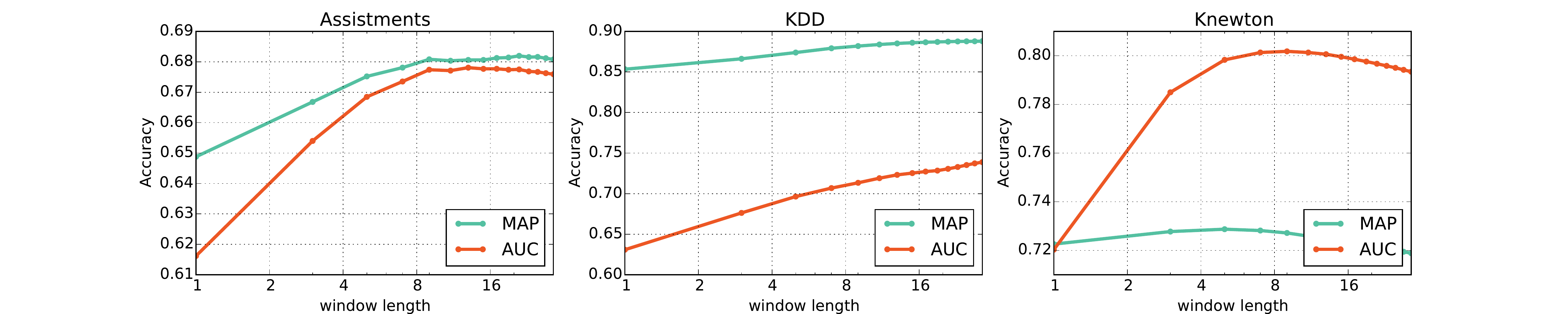}
\caption{Accuracy metrics for the three data sets computed using a rolling window of previous responses, as a function of window length. Response accuracy is computed by predicting correct if the majority of responses in the window are correct.}\label{fig:box_len}
\end{figure*}

\begin{table*}[t!]
\centering
\begin{tabular}{@{}rrrrr@{}}\toprule
            & IRT & HIRT & tIRT & DKT$^\ast$ \\\midrule
ASSISTments & {\tt problem\_id} & {\tt template\_id} $\to$ {\tt problem\_id} & {\tt problem\_id} & {\tt template\_id} \\
KDD & {\tt Step Name} & {\tt Problem Name} $\to$ {\tt Step Name} & {\tt Step  Name} & {\tt KC} \\
Knewton & {\tt item\_id} & {\tt concept\_id} $\to$ {\tt item\_id} & {\tt item\_id} & {\tt concept\_id}\\
\bottomrule
\end{tabular}
\caption{Item labels yielding best results for each model and data set.  For HIRT, the first label specifies the difficulty mean grouping identifier, and the second the item identifier.}
\label{tab:bestlabels}
\end{table*}

\section{Evaluation methodology}\label{sec:evaluation}

\subsection{Parameter Selection} \label{sec:parselect}

For each data set, 20\% of students were first set aside for parameter selection, which we performed as follows:

\begin{itemize}
  \item For 1PO IRT there were no parameters to select.
  \item For HIRT, we swept values of the variances $\tau^2$ and $\sigma^2$ of the group means and item difficulties respectively, including regimes ($\tau^2$ small) which made the model mathematically equivalent to 1PO IRT. 
  \item For TIRT, we swept the temporal smoothness parameter $\gamma^2$, including the regime ($\gamma^2$ small) which made the model mathematically equivalent to 1PO IRT.
  \item For DKT, we swept the compression dimension $C$ (the dimension of the space to which the input was projected using a random matrix), the hidden dimension $H$, the dropout probability $p$, and the step size of our gradient ascent.
\end{itemize}

\subsection{Online prediction accuracy} \label{sec:onlinevalidation}

We use an evaluation method we call {\em online response prediction} which matches that of \cite{DKTNIPS}. Students are first split into training and testing populations. Each model is first trained on the training population and the model parameters that are not student-level (item parameters for IRT-based models, weights for neural networks) are frozen. Then for each time $t > 1$ in each testing student's history, we train the student-level parameters in the model on the first $t - 1$ interactions of the student history and allow it to compute the probability that the $t$'th response is correct. This process mirrors the practical task that must be completed by an ITS.

We report two different metrics for comparing the predicted correctness probabilities with the observed correctness values. Accuracy (Acc) is computed as the percent of responses in which the correctness coincides with the probability being greater than 50\%. AUC is the Area Under the ROC Curve of the probability of correctness for each response.

We use five-fold cross validation (by partitioning the students) on the 80\% of the data set remaining after parameter selection (Section~\ref{sec:parselect}), averaging the Acc and AUC metrics over five different splits of the student population.

\section{Results and Discussion}\label{sec:results}

% Display the results and state the main take-away

Table~\ref{tab:bestlabels} enumerates the fields chosen in each data set to identify items and item groups (for HIRT only) that yielded the computationally tractable model with the best results. Note that for the IRT-based models, our validation scheme (Section~\ref{sec:onlinevalidation}) estimates a single number $\theta_{st}$ for each student at each point $t > 1$ of the validation.For computational reasons, it was not feasible to evaluate DKT on fine-grained labels in KDD and Knewton (for ASSISTments, fine-grained labels were tractable but yielded worse results), whereas all IRT variants were able to process data at the finest levels.

We trained and validated each of the three models on each of the three data sets as described in Sec.~\ref{sec:evaluation}. The results on our evaluation task are summarized in Figure~\ref{fig:results}. The results clearly indicate that simple IRT-based models do as well or significantly better than DKT across all data sets.

% What are some potential explanations for the results?

% The hierarchical structure argument
The fact that HIRT is the best-performing model across the board (except for MAP accuracy on the Knewton dataset where TIRT slightly outperforms it) suggests that grouping structure is useful information to exploit when predicting student responses. Indeed, the HIRT model does have access to strictly more information than the other models in that it has both the item and group identifier associated with each interaction. While the DKT model does have the ability to infer item relationships from data, our results indicate that building in this knowledge is more advantageous in a variety of educational settings. One potential area to explore is in learning a hierarchical model purely from the data, which could profit from the structured Bayesian framework without requiring prior information or expert labels.

% The temporal structure argument

The temporal IRT model yielded higher accuracy on the Knewton dataset, but not on the other two data sets.  To understand these effects, we investigated the degree to which temporal structure in the data affects predictive performance by looking at how a naive ``windowed percent correct'' (predict the student will answer the $t$th question correctly if they answer at least half of the previous $w$ questions correctly) model performs as a function of window length $w$ (Figure~\ref{fig:box_len}). The Knewton data set has a clear optimal window length -- integrating over windows too short or too long degraded performance, which is indicative of nontrivial temporal structure. However, for the ASSISTments and KDD data sets, longer window lengths perform equal or better than shorter window lengths, suggesting that static models would do just as well in these cases. Indeed, this would explain why TIRT does more or less the same as baseline 1PO IRT on ASSISTments and KDD but shows significant improvement on the Knewton data set. However, it does not explain why DKT lags regardless of the amount of temporal structure.

% Explain discrepancy with \cite{DKTNIPS}

Finally, we note that our DKT results in Figure~\ref{fig:results} contradict those of \cite{DKTNIPS} on the ASSISTments data set, which reported an AUC of 0.86. We believe this is due to data cleaning issues, specifically the issue of removing duplicates so as not to artificially boost online prediction accuracy, as discussed in Section~\ref{sec:assistments}. Indeed, we were able to reproduce the performance reported in \cite{DKTNIPS} when applying our RNN implementation on the raw data set (with duplicates left in).

Other recent work \cite{Khajah2016deep} points out that the specific method of computing AUC in \cite{DKTNIPS} also significantly affects the reported performance relative to BKT-based models, and further demonstrates that BKT-based models can perform just as well as DKT on a variety of data sets.

\section{Conclusion}\label{sec:conclusion}

% Summarize results

Our results indicate that simple IRT-based models equal or outperform DKT on a variety of data sets, suggesting that incorporating domain knowledge into structured Bayesian models comprises a promising area of future research for modeling student interaction data. 

In our experience, structured models were easier to train and required less parameter tuning than DKT. Moreover, the computational demands of DKT hampered our ability to fully explore the parameter space, and we found that computation time and memory load were prohibitive when training on tens of thousands of items. These issues could not be mitigated by reducing dimensionality without significantly impairing performance. Further work on discriminative models is necessary to bridge this gap, but currently, IRT-based models seem superior both in terms of performance and ease of use, making them suitable candidates for real-world applications (e.g. intelligent tutoring systems, recommendation systems, or student analytics).

% Perhaps combining the two is a new way forward

A promising avenue of research could explore combining the advantages of structured Bayesian models with those of large-scale discriminative models, which have provided superior performance in several other domains, particularly in the large-data regime. A crucial challenge for structured models is how to accommodate the diversity of educational settings from which the data are collected (different content, different classroom environments, etc.) while retaining the structure that drives predictive power and interpretability. 

\small
\bibliographystyle{acm}
\bibliography{main}

\end{document}